\documentclass[letterpaper, 10 pt, journal, twoside]{IEEEtran}
\usepackage{cite}
\usepackage{array}
\usepackage{float}
\usepackage{xcolor}
\usepackage{lipsum}
\usepackage{caption}
\usepackage{wrapfig}
\usepackage{graphicx}
\usepackage{textcomp}
\usepackage{fixltx2e}
\usepackage{hyperref}
\usepackage{subcaption}
\usepackage{algorithmic}
\usepackage{multirow}
\usepackage[euler]{textgreek}
\usepackage[affil-it]{authblk}
\usepackage{amsmath,amssymb,amsfonts}

\title{Fine Robotic Manipulation without \\Force/Torque Sensor\\}

\author{Shilin Shan$^{1,*}$, and Quang-Cuong Pham$^{2}$ 
\thanks{Manuscript received: July 21, 2023; Revised September 26, 2023; Accepted
December 06, 2023.}
\thanks{This paper was recommended for publication by Editor Hyungpil Moon upon
evaluation of the Associate Editor and Reviewers' comments.}
\thanks{$^{1}$ Shilin Shan is with School of Mechanical and Aerospace
Engineering, Nanyang Technological University, Singapore. {\tt\footnotesize
e-mail: shilin.shan153@gmail.com}}
\thanks{$^{2}$ Quang-Cuong Pham is with Eureka Robotics and Singapore Centre for
3D Printing (SC3DP), School of Mechanical and Aerospace Engineering, Nanyang
Technological University, Singapore. {\tt\footnotesize e-mail:
cuong@ntu.edu.sg}}}

\begin{document}

\maketitle

\vspace*{-25pt}
\begin{abstract}

    Force Sensing and Force Control are essential to many industrial
    applications. Typically, a 6-axis force/torque (F/T) sensor is installed
    between the robot's wrist and the end effector to measure the forces and
    torques exerted by the environment on the robot (the external wrench). While
    a typical 6-axis F/T sensor can offer highly accurate measurements, it is
    expensive and vulnerable to drift and external impacts. Existing methods
    aiming at estimating the external wrench using only the robot's internal
    signals are limited in scope. For instance, the estimation accuracy has
    mainly been validated in free-space motions and simple contacts, rather than
    tasks like assembly that require high-precision force control. In this
    paper, we present a Neural-Network-based solution to overcome these
    challenges. We offer a detailed discussion on model structure, training data
    categorization and collection, as well as fine-tuning strategies. These
    steps enable precise and reliable wrench estimations across a variety of
    scenarios. As an illustration, we demonstrate a pin insertion experiment
    with a 100-micron clearance and a hand-guiding experiment, both performed
    without external F/T sensors or joint torque sensors.

\end{abstract}


\section{Introduction}

\IEEEPARstart{F}{orce} sensing and force control are essential in various industrial
applications, from contact-based inspection to assembly, sanding, deburring, and
polishing \cite{siciliano2008springer, suarez2018can, pham2020convex}.
Typically, a 6-axis force/torque (F/T) sensor is installed between the robot's
wrist and the end-effector to measure forces and torques (the external wrench)
applied by the environment to the robot. Although a standard 6-axis F/T sensor
offers highly accurate measurements, it is expensive and vulnerable to drift and
external impacts. Consequently, there has been substantial research geared
towards estimating the external wrench using solely the robot's internal
signals, such as joint position, joint velocity, or motor current readings.

To that aim, there are two main approaches in the literature: model-based and
model-free. The model-based approach constructs parameterized models of the
robot's dynamics, which are then identified using standard parameter
identification techniques \cite{de2005sensorless, van2011estimating,
wahrburg2017motor, 9382115}. However, the precise modeling and identification of
highly nonlinear and nonsmooth phenomena, such as hysteresis and joint friction
in low-speed areas, pose considerable challenges. To address these and simplify
the modeling process, model-free methods based, e.g., on Gaussian Process
Regression and Neural Networks (NN), have been developed in
\cite{nguyen2009model, liu2018end, sharkawy2020human}. Nevertheless, these
methods have a limited scope. Specifically, the accuracy of wrench estimation
has predominantly been validated in scenarios involving only free-space motions
and straightforward contacts. It has not been extensively tested in complex
industrial tasks demanding high-precision force control.

Here, we present a NN-based method and argue that the aforementioned challenges
can be addressed by specifically focusing on the training data and NN structure.
We highlight the importance of collecting and classifying data for both
free-space and in-contact motions in the low-speed scenarios, while also
selecting a model that balances simplicity and accuracy. The main contributions
can be summarized as follows:

\begin{figure}[t]
\vspace*{-10pt}
\centering
    \minipage{0.285\textwidth}
        \includegraphics[width=\linewidth]{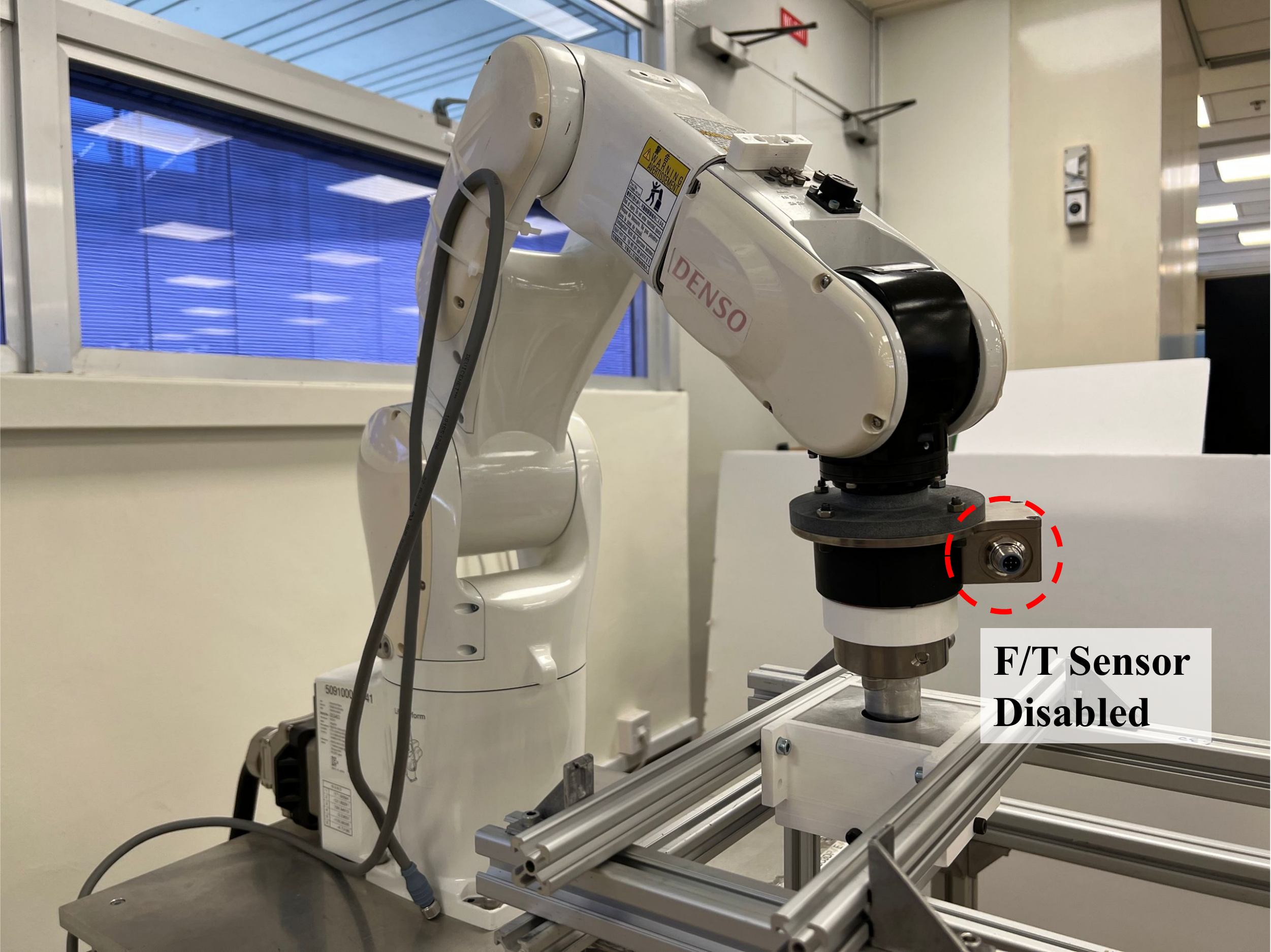}
    \endminipage\hfill
    \minipage{0.196\textwidth}
        \centering
        \includegraphics[width=\linewidth]{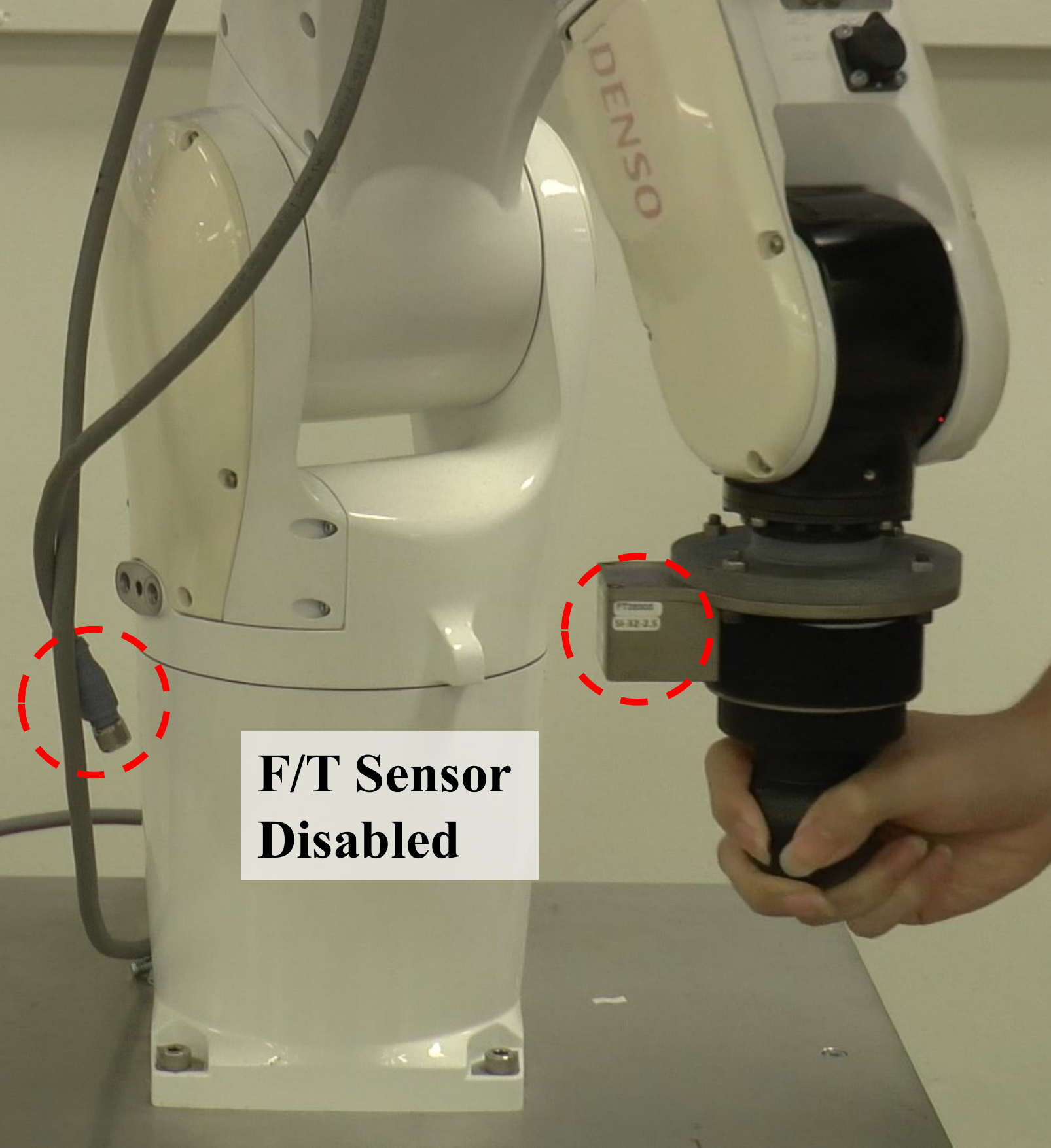}
    \endminipage\hfill
    \caption[Pin insertion and hand-guiding snapshot]{Snapshot of the sensorless
    tight pin insertion and hand-guiding experiment setup. Video demonstrations
    of the experiments in Section VI is available in the supplementary materials
    or at: \href{https://youtu.be/spztx3GzPzc}{https://youtu.be/spztx3GzPzc}}
    \label{fig:snap}
    \vspace*{-10pt}
\end{figure}

\begin{itemize}

\item We investigated systematically the estimation accuracy and computational
cost of various NN structures. The findings offer valuable insights into
selecting the optimal model structure for wrench estimation and hopefully
similar regression tasks across different applications.

\item  We introduce a unique training strategy that initially trains a base
model using unbiased data, followed by fine-tuning to enhance performance in
contact-rich applications. The training data were collected through a
combination of free-space motion planning, high-precision force control, and
physical human-robot interaction tasks.
\end{itemize}

\noindent We demonstrate a pin insertion experiment with 100-micron clearance
and a hand-guiding experiment, both conducted without the use of F/T sensors
during runtime. Notably, an F/T sensor is still required for the one-time data
collection during the training phase. For expansive industrial applications, a
well-trained Neural Network can be ported to other manipulators of the same
model with suitable transfer learning strategies \cite{kim2021transferable}.

The structure of the paper is as follows: Section II offers an overview of
literature related to both model-based and model-free wrench estimation. In
Section III, we introduce the implementation of the model-based scheme and NN
structures. Section IV elaborates on data collection methodologies and provides
a detailed procedure. Section V presents the model training outcomes and delves
into the principles of model selection, considering dataset coverage. Section VI
showcases four experiments performed in both free-space and in-contact
situations. Fig. \ref{fig:snap} provides a snapshot from the sensorless pin
insertion and hand-guiding experiments.

\section{Related Work}

Early research introduced a systematic approach to dynamics identification and
trajectory excitation \cite{khalil2002modeling}. Following this, the renowned
Generalized Momentum Observer (GMO) \cite{de2003actuator, de2005sensorless} was
proposed, aiming to derive high-quality torque residual estimates based on
either driving torque or current measurements. While mathematical analyses offer
crucial insights into rigid body dynamics, approximating joint frictions in
low-speed scenarios is challenging. These scenarios might encompass highly
nonlinear phenomena, such as hysteresis \cite{liu2015experimental}. Beyond the
conventional friction models addressing Coulomb friction, stiction, viscous
friction, and Stribeck friction \cite{liu2021sensorless}, alternative methods
have been suggested. These methods aim to mitigate friction errors by employing
sigmoid functions \cite{gaz2018model, gaz2019dynamic} or the Kalman Filter
\cite{wahrburg2017motor}. Recent studies have also put forth semi-parametric
approaches that utilize Neural Network models to learn joint motor friction
\cite{liu2018end} and counteract non-modeled effects \cite{hu2017contact}.
Despite the in-depth error analyses provided, the robustness of these methods in
high-precision practical applications remains to be ascertained.

Another approach is to bypass manual selection of the entire model structure and
instead allow the model to learn from one-time data collection. Early research
has explored non-parametric regression-based approaches for model identification
\cite{schaal2002scalable, vijayakumar1997local}. Building upon these methods,
several studies have demonstrated the ease of training non-parametric
learning-based approaches such as Gaussian Process Regression (GPR) and Locally
Weighted Projection Regression (LWPR) \cite{nguyen2008local, nguyen2009model,
gijsberts2013real}. By avoiding manual selection of the model structure,
regression-based approaches eliminate human bias and enable the model to learn
the optimal structure given simple hyperparameters. GPR, which has been trained
and validated using hand-guiding experiments \cite{nguyen2009model}, exhibits
high accuracy in trajectory tracking. However, this task alone may not
adequately represent contact-rich industrial tasks. Additionally, as discussed
in Section III, the hand-guiding dataset may be biased due to the strong
correlation between end-effector force and motion.

Following Section I, we propose the use of Neural Networks to avoid complex
mathematical modeling while ensuring estimation accuracy. Neural Networks have
proven to be effective in approximating nonlinear mappings in various domains.
Recent research has applied different Neural Network structures to various
industrial tasks. For example, Multilayer Perceptron (MLP) has been proposed in
\cite{yilmaz2020neural, sharkawy2020human, kim2021transferable} for Inverse
Dynamics Identification and contact detection by classifying the robot motion
status. Convolutional Neural Networks (CNN) have been utilized in
\cite{lee2018interaction, xia2021sensorless} to estimate external forces by
detecting the physical deformation of objects. Recurrent Neural Networks (RNN),
particularly Long Short-term Memory Networks (LSTM), have been proposed in
\cite{farazi2017online, wu2021hysteresis} for robot tracking and hysteresis
approximation. In general, MLP provides a direct nonlinear mapping from the
input to the output. CNN and LSTM, on the other hand, are specialized in
extracting specific features from input data and capturing temporal information
\cite{goodfellow2016deep}, respectively. 

In this paper, we evaluate the performance of MLP, CNN, and LSTM in the wrench
estimation task and demonstrate that MLP can achieve more accurate estimation
given equal inference time. In Section VI, we compare the fine-tuned MLP with
the model-based method - implemented following standard dynamics identification
procedure \cite{khalil2002modeling}, the GMO scheme proposed in
\cite{gaz2018model}, and the friction compensation technics discussed in
\cite{wahrburg2017motor, gaz2019dynamic}.

\section{Preliminaries}

\subsection{Model-based Identification}

Consider a robot with rigid joint, the robot dynamics can be described as follows:
\vspace{-1pt}
\begin{equation}
\label{e1}
\vspace{-1pt}
M(q)\ddot{q} + C(q,\dot{q})\dot{q} + g(q) + \tau_f(\dot{q}) = \tau_{motor} + \tau_{ext}
\end{equation}
where $M(q)$, $g(q)$, $\tau_f(\dot{q})$, $\tau_{motor}$, $\tau_{ext}$ are the
inertia matrix, Gravity vector, joint friction, motor driving torque, and
external torque, respectively. $C(q,\dot{q})$ is the factorized Coriolis matrix
such that $\dot{M}-2C$ is skew-symmetric. Such a property enables the design of
a Generalized Momentum Observer (GMO) \cite{de2005sensorless} that avoids the
use of noisy acceleration, being described by the following equation:
\vspace{-3pt}
\begin{equation}
    \vspace{-3pt}
    \label{e2}
    r(t) = L\left(M(q)\dot{q}-\int_{0}^{t}(\bar{\tau}+r)ds\right)
\end{equation}
where $r$ is the approximated residual, $L$ is the diagonal gain matrix, and
$\bar{\tau} = C^T(q, \dot{q}) \dot{q} - g(q) - \tau_f(\dot{q}) + \tau_{motor}$.
Considering a setting where only current measurements are available, dynamics
parameter identification can be conducted on the current level using the
following equation:
\vspace{-3pt}
\begin{equation}
    \vspace{-3pt}
    \label{e3}
    Y(q,\dot{q},\ddot{q})\hat{\pi}=I
\end{equation}
where $Y$, $\hat{\pi}$, $I$ are the regressor matrix, dynamics parameters, and
motor current vector, respectively. The exciting trajectory proposed in
\cite{gaz2019dynamic} was considered for data collection. A standard
pseudoinversion was used for identification: $\hat{\pi} = \bar{Y}^\# \bar{I}$,
where $\bar{Y}$ is the stacked regressor matrix evaluated with the collected
data, and $\bar{I}$ is the stacked current measurements.

The end-effector wrench can be approximated using:
\vspace{-3pt}
\begin{equation}
    \vspace{-3pt}
    F = (J^T)^\#KI
\end{equation}
where $F$, $J$, $K$ denote the external wrench, the geometric Jacobian and the
diagonal motor constant matrix. However, the motor constants are not available
for our robot model. Supplementary experiment was conducted, calculating
external joint torques with wrench measurements, thereby approximating the motor
constant with the estimated current residuals from equation (\ref{e2}). 

Notably, we used a friction model discussed in \cite{wahrburg2017motor}:
\vspace{-3pt}
\begin{equation}
    \label{e4}
    \tau_f(\dot{q_i}) = \left(\Bigl(C_{C,i} + C_{S,i} \cdot 
    e^{-\bigl(\frac{\dot{q_i}^2}{v_{S,i}}\bigr)^2}\Bigr) sgn(\dot{q_i})
                            + C_{V,i} \cdot \dot{q_i} \right)
\end{equation}
where $i$ denotes the $i^{th}$ joint, $C_{C,i}$, $C_{S,i}$, $C_{V,i}$, and
$v_{S,i}$ are the coefficient for Coulomb friction, stiction, viscous friction
and Stribeck friction, respectively. Identified large Coulomb coefficients can
induce high-frequency oscillations in estimations when $\dot{q}$ approaches
zero. To enable a smooth transition around this zero velocity, a Sigmoid
function is employed \cite{gaz2019dynamic}:
\begin{equation}
    \tau_{f_C,i} = \frac{\varphi_{1,i}}{1+e^{-\varphi_{2,i}(\dot{q_i}+\varphi_{3,i})}} -
                   \frac{\varphi_{1,i}}{1+e^{-\varphi_{2,i}\varphi_{3,i}}}
\end{equation}
where $\tau_{f_C,i}$ can replace the $C_{C,i} \cdot sgn(\dot{q_i})$ term in
equation (\ref{e4}), and $\varphi_{1,i}$, $\varphi_{2,i}$, and $\varphi_{3,i}$
are the parameters to be identified. Despite handling friction terms carefully
in the low-speed scenario, this implementation still shows inaccuracies in the
applications discussed in Section VI, especially in static scenarios where
hysteresis is observed to be a prominent issue.
\vspace*{-11pt}

\subsection{Neural Networks Model Structure}

\begin{figure}[htbp]
    \centering
        \minipage{0.22\textwidth}
            \includegraphics[width=\linewidth]{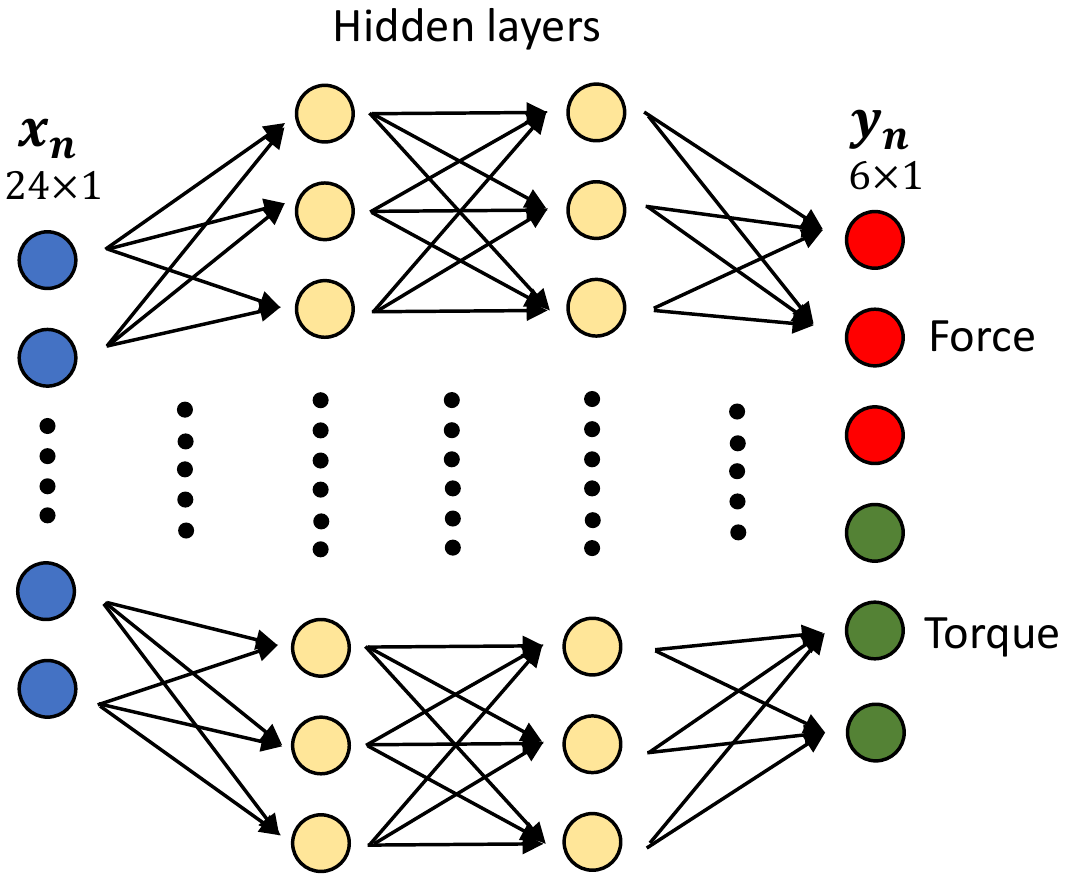}
        \endminipage\hfill
        \minipage{0.26\textwidth}
            \centering
            \includegraphics[width=\linewidth]{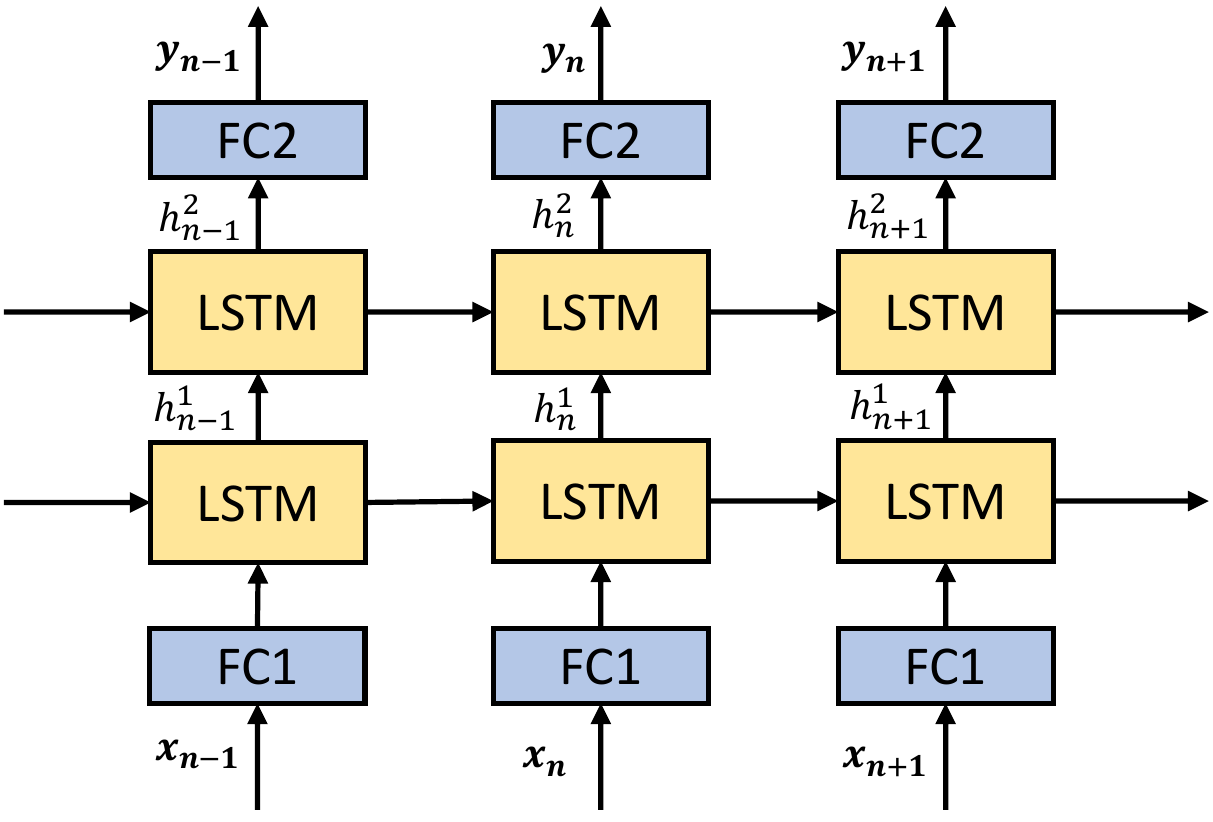}
        \endminipage\hfill
        \vspace{2pt}
        \minipage{0.48\textwidth}
            \centering
            \includegraphics[width=\linewidth]{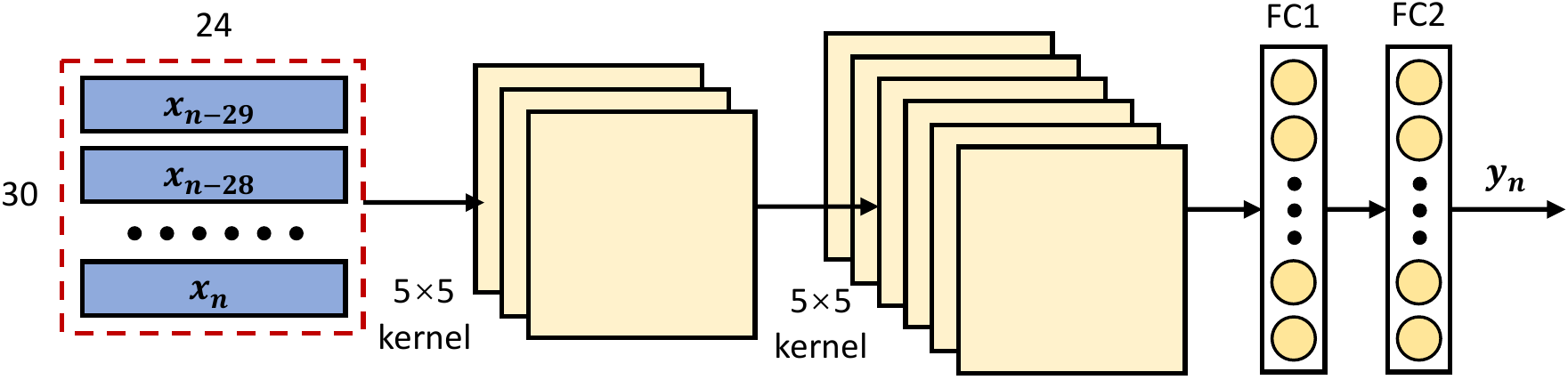}
        \endminipage\hfill
        \caption{The model structure of MLP (top left), LSTM (top right) and CNN
        (bottom) implemented for wrench estimation}
        \vspace*{-8pt}
        \label{Fig:model_structures}
\end{figure}

The model structures of MLP, LSTM, and CNN implemented and used for comparisons
are shown in Fig. \ref{Fig:model_structures}. 

\subsubsection{MLP}
The model takes each frame of joint currents and states, including position,
velocity, and acceleration, as input and maps them to the end-effector wrench.
The training time, inference time, and estimation accuracy are highly influenced
by the network depth and hidden layer size. The relationship between these
hyperparameters and the coverage of the training set will be discussed in
Section V.

\subsubsection{LSTM}
The recurrent structure of the model comprises one fully-connected (FC) layer,
two LSTM layers, and another FC layer. It takes in one frame of joint currents
and states and estimates the wrench for the same frame. The latent information,
represented by the hidden states $h_n$ and cell states $c_n$, is then propagated
to the subsequent LSTM layers for the next frame. This process enables the model
to learn from both long-term and short-term joint state memories
\cite{farazi2017online, wu2021hysteresis}. Further information regarding the
related theory and the detailed structure of the LSTM layer can be found in
\cite{hochreiter1997long}.

\subsubsection{CNN}
The model typically receives 2D images as input, enabling it to understand the
pixel relationships and extract relevant features. In the case of joint currents
and states as input, the data pattern is adapted to a CNN by concatenating the
current data frame with multiple previous frames, forming a 2D matrix. The
convolutional kernel is designed to extract features from individual input
frames as well as across multiple frames. A similar idea can be found in
\cite{zhang2022robot}, where CNN was used for identifying the features across
dynamics equation and friction components. Additional information about the
related theory and convolution process can be found in
\cite{lecun1995convolutional}.

\subsubsection{Training Framework}
We implemented the models and data loader, then trained the models with PyTorch.
The loss function and optimization method are MSEloss and Adam Optimizer,
respectively. ReLU activation was applied to all FC and convolutional layers,
while Sigmoid and tanh were used as the default activation functions for LSTM
layers. Detailed information about the hyperparameters will be discussed in
Section V along with the training data.

\section{Training Data Generation}

In this section, we introduce a unique data collection method, grounded in the
concepts of dynamics information bias and the strategy of problem-solving
fine-tuning.

\subsection{Fundamental Concepts}

For the model-based approach, we can adopt the method outlined in
\cite{gaz2019dynamic}, using conventional sinusoid trajectories excited in the
joint space. The well-defined robot dynamics enables straightforward parameter
fitting with minimal training data, made possible through optimized trajectory
design. Conversely, for the NN-based approach, there are no prior assumptions,
meaning the learning result depends entirely on the training data distribution.
In response, we introduce a data collection scheme designed to encompass a broad
range of scenarios. Specifically, we segment the training data into two
categories: a base dataset and a fine-tuning dataset.

The base datatset was collected while the robot end-effector followed
pre-planned trajectories, namely Free-Space DataSet (FSDS). FSDS was designed to
train the Neural Network for foundational force/torque sensing ability, and
therefore, it includes random manual contacts measured by an F/T sensor at the
end-effector throughout the data collection process. FSDS serves as the backbone
of the training set, as it provides the model with unbiased information.
Specifically, the robot's motion is independent of the wrench, ensuring
comprehensive coverage of the robot's dynamics.

The fine-tuning dataset should be collected considering potential model
application scenarios. We collected two dataset, namely Contact Sliding DataSet
(CSDS) and Hand-Guiding DataSet (HGDS), to optimize model performance for these
particular tasks. In contrast to the base dataset, the fine-tuning datasets
generally exhibit two characteristics: (i) The trajectories are not pre-planned
smooth trajectories but real-time evaluated trajectories arising from control
laws, which often include noisy measurements and various uncertainties. (ii) The
robot motions are highly coupled with the external forces. For instance,
friction force always resists the direction of sliding in hybrid force control
tasks, and the end-effector fully complies with the external wrench during
hand-guiding.

Although our experiments indicate that the model fine-tuned with one task, e.g.,
hand-guiding, can effectively be applied to another task, e.g., contact sliding,
with considerable accuracy, we aim to mitigate any potential bias by identifying
the skewed information.
\vspace*{-5pt}

\subsection{Training Data Collection}

The data collection and experiments presented in the following sections were
conducted using the Denso-VS060, a 6-axis position-controlled industrial robot.
During FSDS and HGDS collection, a 3D-printed sphere with a diameter of 50mm was
used as the end-effector. For CSDS collection, an aluminum cylinder pin with a
diameter of 20mm was used. Both end-effectors were mounted on the F/T sensor.

\subsubsection{Free-space DataSet (FSDS)}

The collection procedure commenced by sampling multiple random end-effector
positions within the workspace. Each position was assigned random roll, pitch,
and yaw angles selected within the specified ranges. A trajectory planner then
determined the shortest trajectory between each point sequentially. During the
execution of these pre-planned trajectories, forces were manually and
continuously applied to the end-effector in a random manner. Given that Denso
VS060 provides only joint position measurements, we obtained joint velocity and
acceleration through the first and second derivatives of joint position. To
marginally reduce signal noise, we employed a third-order Butterworth Filter,
yet some noise was retained, positing that NN models should train on noisy
signals for improved resilience. The training data, which incorporates F/T
sensor readings, joint current, position, velocity, and acceleration, was
recorded for FSDS and all subsequent datasets under 100Hz system frequency.

\subsubsection{Contact Sliding DataSet (CSDS)}

CSDS was generated through direct end-effector contacts with a steel plate,
which was secured within the FSDS workspace. Multiple contact points were
randomly sampled in the specified area. For each contact point, the end-effector
repeated the following: (i) Making contact with the plate at a random angle and
applying a desired force for 30 seconds. To induce disturbances, a random
reference force was sampled within the range of 4N to 30N every 0.2 second. (ii)
Sliding towards the next contact point while maintaining the applied force.
(iii) Disengaging from the plate after reaching the next contact point.

Fig. \ref{Fig:blockdiagram_all} illustrates that the contact force was
controlled through an admittance control scheme in the workspace. Sliding
motions parallel to the plate followed the pre-planned straight trajectories
between points, and therefore, sliding friction was not actively controlled.
Using the Inverse Kinematics Model, we can evaluate the commanded joint
positions from the desired X, Y, and Z coordinates.

The CSDS collection should be repeated for multiple contact planes in order to
cover a possibly large portion of the workspace. However, achieving this may
require a flexible experimental setup that allows for adjustable plane locations
and orientations. The ideal setup for full-range data collection would involve
two robots pushing against each other to simulate in-contact effects.
Alternatively, the contact plate could be mounted to the robot's end-effector,
enabling programmable plane locations and trajectories. Due to the constraints
of the setup, we collect CSDS data on fixed contact planes.

\subsubsection{Hand-Guiding DataSet (HGDS)}

HGDS was generated when the robot executed a hand-guiding task, where the
end-effector responded compliantly to external force measurements. The same
admittance control law in Fig. \ref{Fig:blockdiagram_all} was utilized but
applicable to the three-dimensional workspace.

\subsection{Training Data Visualization}

\begin{figure}[t]
    \vspace{3pt}
    \centering
    \includegraphics[width=0.48\textwidth]{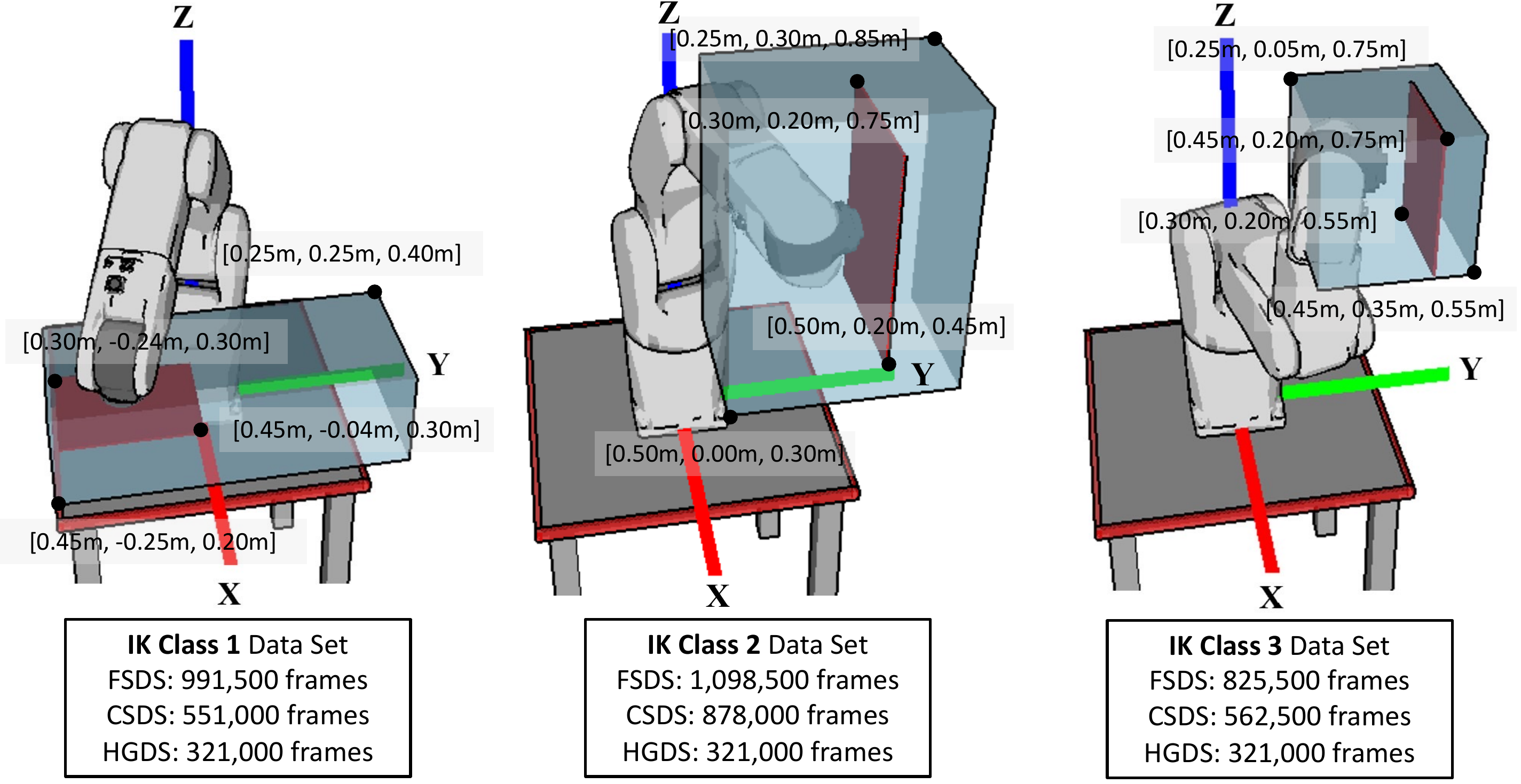}
    \caption{Visualization of the data region for three workspaces with three
    different IK Classes. The transparent blue box and opaque red plane indicate
    the bounding boxes and location of the fixed plates, respectively. The
    postures of Denso show the center of the nearest joint position clusters, or
    IK solutions, based on which the data was collected.}
    \label{Fig:data_ik}
    \vspace{-5pt}
\end{figure}

\begin{figure}[t]
    \vspace*{3pt}
    \centering
        \begin{subfigure}[a]{0.47\textwidth}
            \centering
            \includegraphics[width=\textwidth]{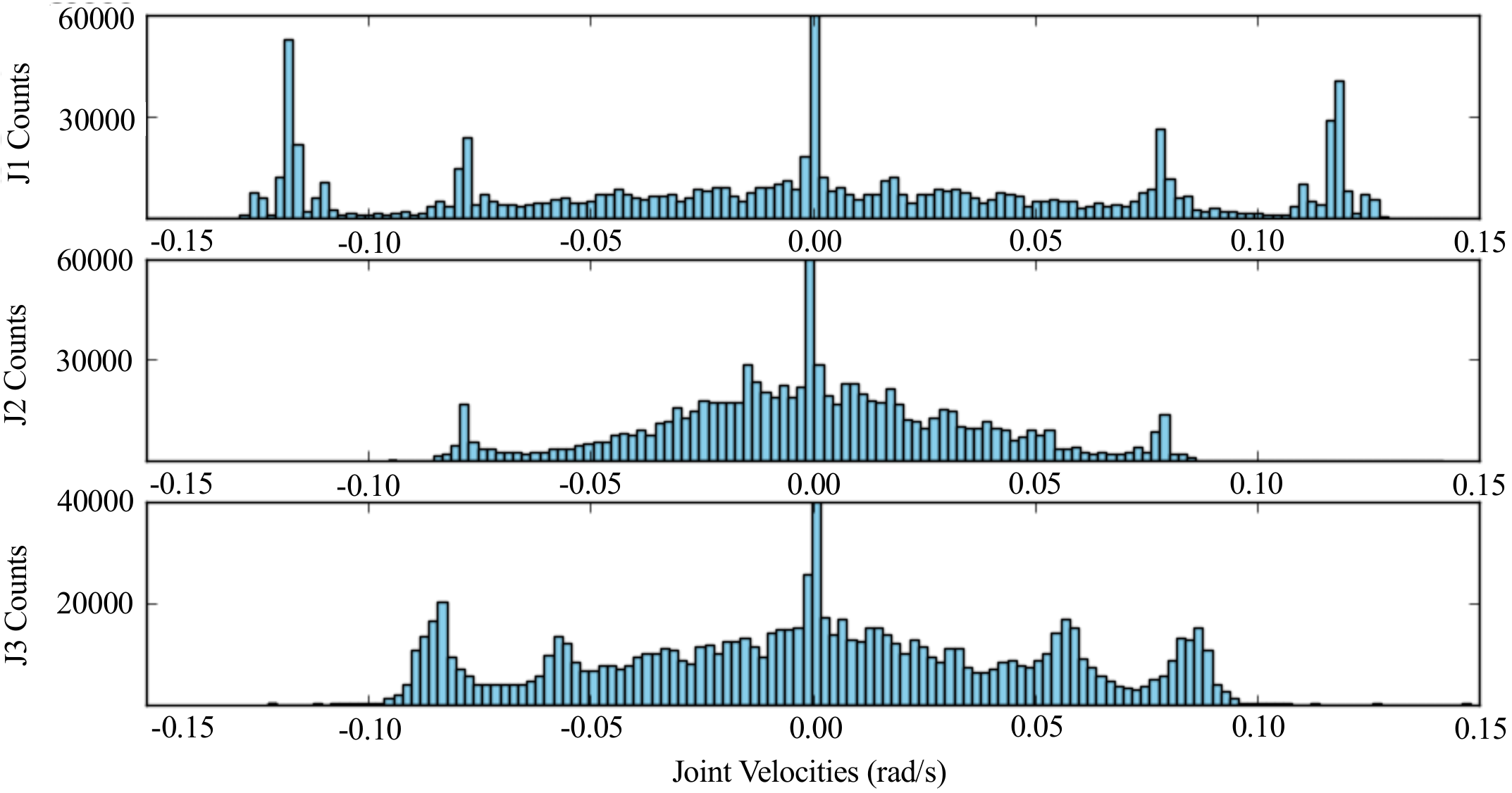}
            \vspace{-18pt}
            \caption{}
            \label{fig:vel_dis_free}
            \vspace{-1pt}
        \end{subfigure}
        \begin{subfigure}[b]{0.47\textwidth}
            \centering
            \includegraphics[width=\textwidth]{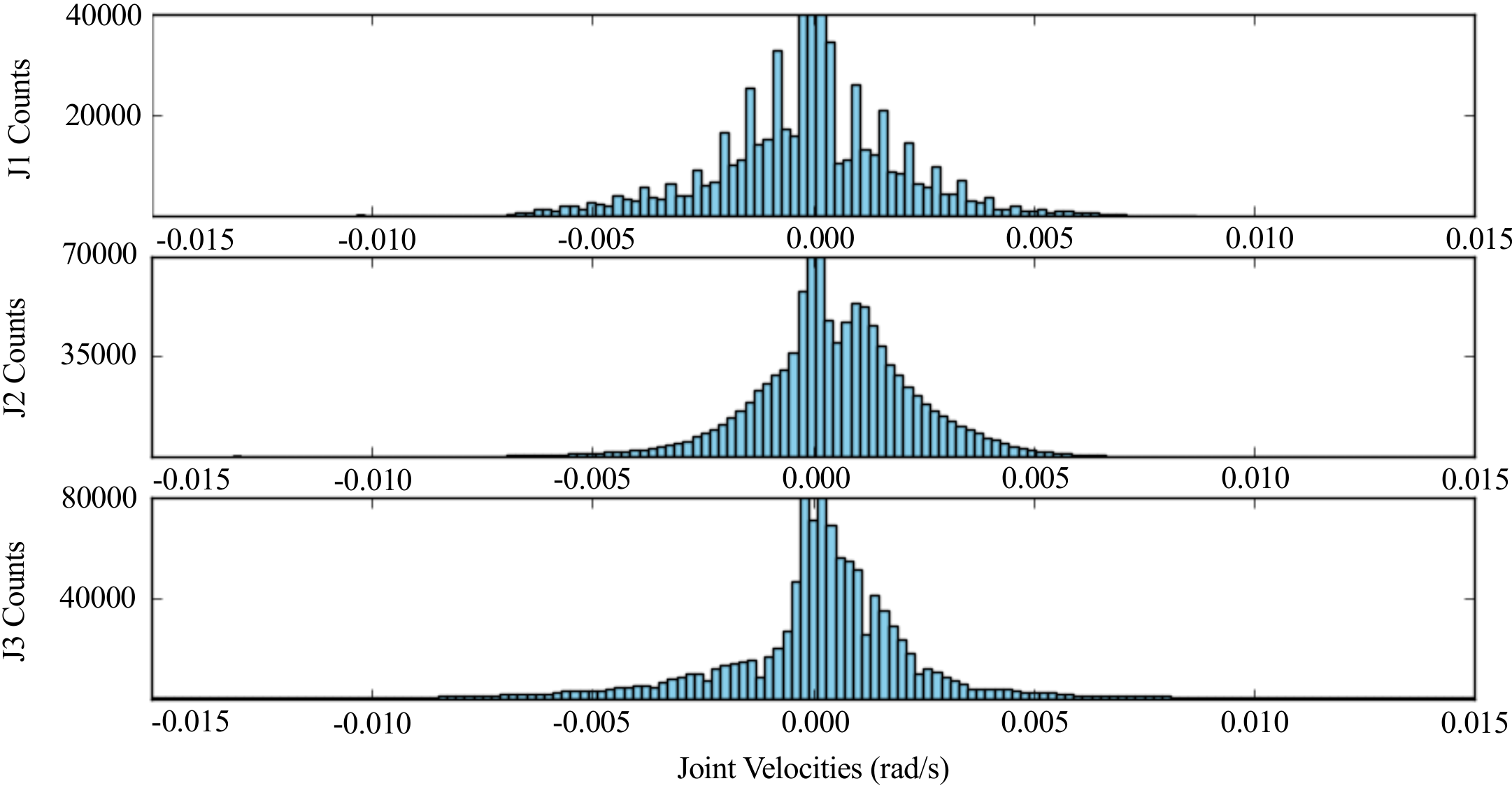}
            \vspace{-18pt}
            \caption{}
            \label{fig:vel_dis_contact}
        \end{subfigure}
        \vspace{-5pt}
        \caption{Velocity distribution of Joint 1, 2, and 3 in (a) FSDS; (b)
        CSDS shown in histograms.}
        \label{Fig:vel_dis}
        \vspace{-10pt}
\end{figure}

We collected the training data considering three dissimilar Inverse Kinematics
Classes (IK Classes). each IK Class incorporates their respective FSDS, CSDS,
and HGDS. Fig. \ref{Fig:data_ik} provides a visualization of the IK Classes'
postures and the corresponding workspace regions in an OpenRave simulated
environment. The figure also shows the Cartesian coordinates of the workspace,
contact plane, as well as the total number of data frames used for training. The
velocity distribution of the training data is shown in Fig. \ref{Fig:vel_dis},
suggesting that all datasets were collected under low-speed scenarios. Due to
the limited space, we show only the distribution of Joint 1, 2, and 3 in FSDS
and CSDS. The remaining datasets show similar characteristics as the data being
displayed.

\section{Model Training and Comparison}

\vspace*{-2pt}
When a lightweight NN model is trained with a large dataset, it tends to
underfit the training data, resulting in inaccurate estimation. To address this
issue, we propose enlarging the hidden layers or increasing the network depth to
capture higher non-linearity and diversities introduced by dissimilar IK
Classes. However, instantaneous wrench feedback is crucial in real-time tasks,
particularly high-precision force-control tasks such as assembly. The challenge
arises when using a large neural network model, as it can lead to longer
inference times. In general industrial application scenarios, additional
computational resources such as GPUs can be costly. Thus, there is a tradeoff
between the model size and inference time that needs to be considered, taking
into account the expected performance of the model in applications.

To understand the relationship between training data coverage and model
behaviors, and to optimize the model size for applications, we experimented with
different model sizes using the FSDS from either a single IK Class or all IK
Classes. Specifically, we trained the proposed MLP, LSTM, and CNN models by
using the following parameters: (i) MLP model with two hidden layers and three
candidate layer sizes: 256, 512, 1024; (ii) LSTM model with FC layers each
comprising 256 neurons, input length of 100, and two candidate hidden state
size: 128, 256; (iii) CNN model with two convolutional layers consisting of 3
and 6 channels, 5x5 convolution kernels, and two FC layers each comprising 512
neurons.

\begin{table}[htbp]
    \vspace*{6pt}
    \centering
    \setlength{\tabcolsep}{5.5pt}
    \begin{tabular}{c c|c c c c c c}
    \hline
    \multicolumn{8}{c}{Test Set RMSE (N for F; Nm for T)} \\ \hline
    Model & Datasets & Fx & Fy & Fz & Tx & Ty & Tz \\ \hline
    CNN        &   1   & 4.09 & 4.52 & 6.80 & 0.30 & 0.31 & 0.19 \\
    CNN        & 1,2,3 & 4.46 & 4.98 & 7.52 & 0.38 & 0.38 & 0.23 \\
    LSTM HS256 &   1   & 2.86 & 2.68 & 3.69 & 0.17 & 0.19 & 0.08 \\
    LSTM HS128 &   1   & 3.15 & 2.95 & 4.04 & 0.19 & 0.21 & 0.08 \\
    LSTM HS256 & 1,2,3 & 3.04 & 2.97 & 4.70 & 0.20 & 0.21 & 0.08 \\
    LSTM HS128 & 1,2,3 & 3.40 & 3.36 & 5.21 & 0.21 & 0.23 & 0.09 \\
    MLP LS1024 &   1   & \textbf{2.48} & \textbf{2.22} & \textbf{3.39} & 
                         \textbf{0.15} & \textbf{0.17} & \textbf{0.06} \\
    MLP LS512  &   1   & 2.90 & 2.61 & 3.64 & 0.17 & 0.19 & 0.07 \\ 
    MLP LS256  &   1   & 3.68 & 3.37 & 4.45 & 0.23 & 0.25 & 0.08 \\ 
    MLP LS1024 & 1,2,3 & 2.60 & 2.44 & 3.47 & 0.17 & 0.18 & 0.07 \\
    MLP LS512  & 1,2,3 & 3.28 & 3.00 & 4.56 & 0.20 & 0.21 & 0.08 \\
    MLP LS256  & 1,2,3 & 4.30 & 3.89 & 5.33 & 0.26 & 0.31 & 0.09 \\ \hline
    \end{tabular}

    \begin{tabular}{c}
        \\
    \end{tabular}

    \begin{tabular}{c|c c c c c}
    \hline 
    \multicolumn{6}{c}{Inference Time} \\ \hline
    & CNN & LSTM 256 & MLP 256 & MLP 512 & MLP 1024 \\ \hline
    Time & 1.95ms & 2.21ms & 0.20ms & 0.88ms & 2.43ms \\ \hline 
    \end{tabular}

    \caption{RMSE on the same test set and model inference time. `LS' and `HS'
    stand for hidden Layer Size for MLP and hidden state size for LSTM.
    'Datasets' suggests the datasets involved in training.}
    \label{Table:test_RMSE}
    \vspace*{-5pt}

\end{table}

The quantized error, evaluated on the test set of 150,000 frames, is shown in
Table \ref{Table:test_RMSE}. As we expand the size of hidden layers, the
estimation accuracy tends to improve across all dimensions. However, this also
leads to an increased inference time. Notably, smaller models shows a
significant decrease in accuracy when encompassing all three IK Classes, whereas
the impact on larger models is less evident. 

The tested CNN model displays a higher RMSE compared to both the MLP and LSTM
models, even with its significant computational cost. The MLP and LSTM models,
having similar inference times, exhibit nearly equivalent RMSE values and
demonstrate consistent drops in accuracy across all IK Classes. We favored the
MLP structure for two main reasons: (i) The recurrent nature of the LSTM can
result in inconsistent estimations, where a steady input may yield drifting
outputs, potentially undermining its reliability; and (ii) The MLP
implementation is more straightforward. In conclusion, for all subsequent
control tasks in our experiments, we utilized an MLP model with 2 hidden layers,
each having 1024 neurons.

\section{Experimental Applications}

\subsection{Wrench Estimation in Free Motion}

\begin{figure}[t]
    \centering
    \includegraphics[width=0.48\textwidth]{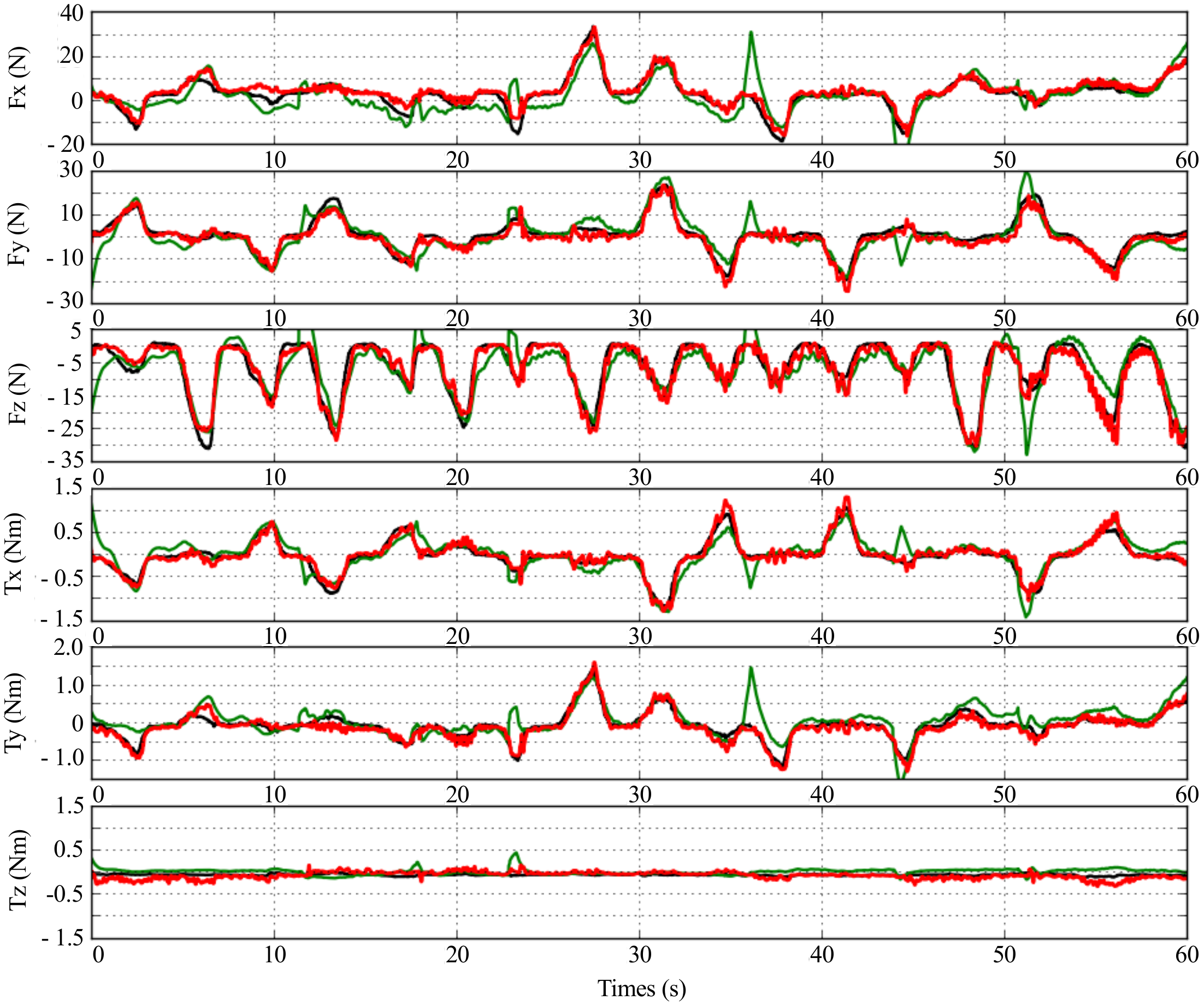}
    \vspace{-5pt}
    \caption{Comparison between the NN-estimated (red) GMO-estimated (green) and
    measured wrench (black) for random pre-planned free-motion trajectories.}
    \label{fig:FM_all}
    \vspace{-5pt}
\end{figure}

\begin{table}[t]
    \vspace*{6pt}
    \centering
    \setlength{\tabcolsep}{5.5pt}
    \begin{tabular}{c | c c c c c c}
    \hline
    \multicolumn{7}{c}{Fig. \ref{fig:FM_all} RMSE (N for F; Nm for T)} \\ \hline
    Model & Fx & Fy & Fz & Tx & Ty & Tz \\ \hline
    Proposed & 2.17 & 2.22 & 2.28 & 0.09 & 0.12 & 0.08 \\
    GMO      & 5.22 & 4.12 & 5.62 & 0.20 & 0.27 & 0.10 \\ \hline
    \end{tabular}
    \caption{RMSE for the free-motion online test.}
    \vspace{-5pt}
    \label{Table:FM_error}
\end{table}

The free-motion experiment served as an online test set to evaluate the
performance of the trained model. Test trajectories were randomly generated and
forces were applied following the same manner as FSDS collection for IK Class 1.
The NN model used in this experiment was trained with only FSDS.

In Fig. \ref{fig:FM_all} and Table \ref{Table:FM_error}, we compare the
estimations from the proposed method, GMO, and sensor measurements. It's evident
that while the implemented GMO could capture the force variation in this
low-speed scenario, its overall accuracy is unsatisfactory. In contrast, the
proposed method consistently delivers promising results, showing both close
alignment with the measurements in the Figure and low RMSE values.

\subsection{Force Control for In-contact Spiral Sliding}

\begin{figure}[t]
    \centering
    \includegraphics[width=0.48\textwidth]{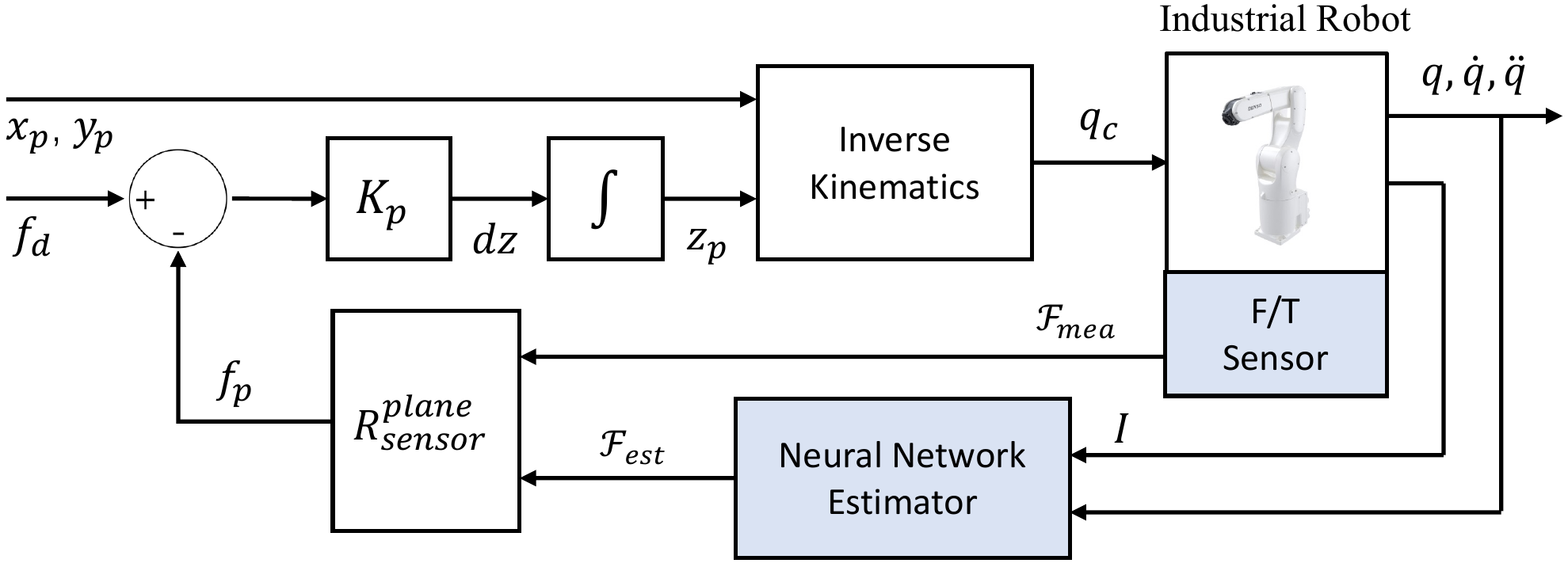}
    \caption{Block diagram for force control on contact planes. $x_p$, $y_p$,
    and $z_p$ are the end-effector coordinate in the plate's frame. $f_d$,
    $f_p$, $I$ and $R^{plane}_{sensor}$ indicate the desired force on the plate,
    the sensed or estimated force, the joint current, and the frame
    transformation matrix.}
    \label{Fig:blockdiagram_all}
    \vspace{-10pt}
\end{figure}

\begin{figure}[t]
    \vspace{2pt}
    \centering
    \includegraphics[width=0.48\textwidth]{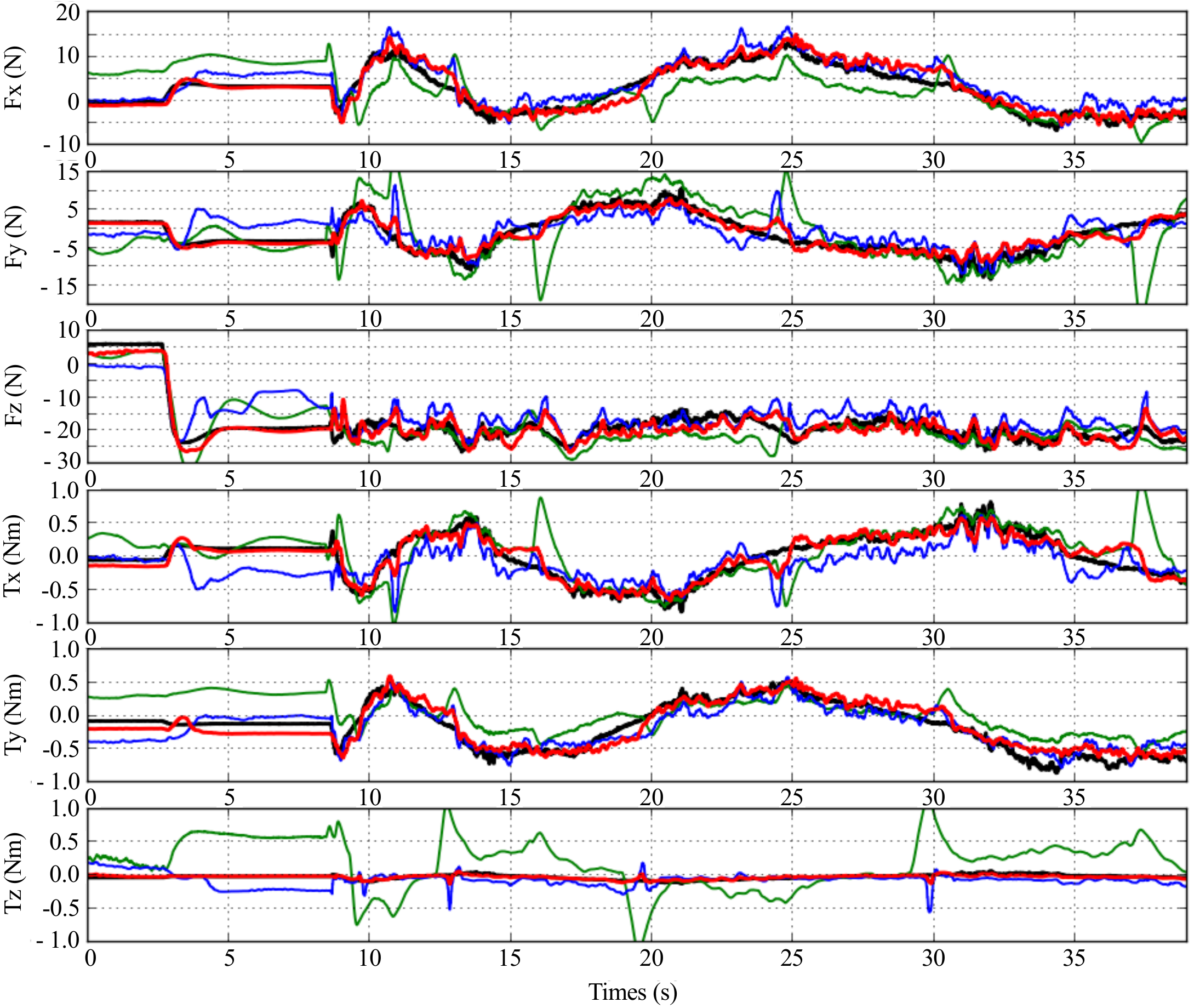}
    \caption{Comparison between the fine-tuned model (red), base model (blue),
    and GMO (green) estimations with the measured wrench (black) for in-contact
    spiral sliding.}
    \label{Fig:SS}
    \vspace{-2pt}
\end{figure}

\begin{table}[t]
    \vspace{4pt}
    \centering
    \setlength{\tabcolsep}{5.5pt}
    \begin{tabular}{c | c c c c c c}
    \hline
    \multicolumn{7}{c}{Fig. \ref{Fig:SS} RMSE (N for F; Nm for T)} \\ \hline
    Model  & Fx & Fy & Fz & Tx & Ty & Tz \\ \hline
    CSDS Fine-tuned  & 1.60 & 1.67 & 2.37 & 0.12 & 0.12 & 0.02 \\
    FSDS Trained     & 2.43 & 3.49 & 4.82 & 0.26 & 0.16 & 0.11 \\
    GMO              & 4.33 & 5.73 & 3.74 & 0.28 & 0.32 & 0.41 \\ \hline
    \end{tabular}
    \caption{RMSE for the in-contact spiral sliding experiment.}
    \vspace{-8pt}
    \label{Table:SS_error}
\end{table}

We employed an in-contact spiral sliding experiment to highlight the benefits of
fine-tuning a model for specific applications. The end-effector traced a
predetermined spiral trajectory on a fixed-position plate. Instead of depending
on the F/T sensor for control feedback, we leveraged the estimation from the NN
model, which was initially trained by FSDS and subsequently fine-tuned by CSDS.
Fig. \ref{Fig:blockdiagram_all} provides the block diagram of the closed-loop
control.

In the experiment, the end-effector initially moved towards the plate, then
applied constant force of 20N at a contact angle of 10 degrees. Subsequently, it
slid along a spiral trajectory once the force was stabilized. Fig. \ref{Fig:SS}
provides a visualization of the experiment and Table \ref{Table:SS_error} shows
the RMSE comparison. The effect of force control can be interpreted from the
Z-axis force. Despite the exclusion of spiral trajectories from the CSDS, the
fine-tuned model still achieved accurate wrench estimation by learning from
simple straight sliding trajectories. 

In contrast, both the base model and GMO show reduced accuracy, particularly
during contact when the robot remains static from 3s to 8s. In the sliding
phase, both methods exhibit significant peak errors, with joint 6 being notably
affected. These inaccuracies can be caused by the robot's extreme low-speed
operation, leading to the prominent hysteresis issue for our robot model. The
same issue also accounts for the large peak errors displayed in Fig.
\ref{fig:FM_all}, taking place during speed inversion. Addressing these
hysteresis errors often requires a non-local memory of robot states
\cite{liu2015experimental}. The enhanced performance of the fine-tuned model in
tackling this issue can be attributed to two primary factors: (i) CSDS is a
dataset collected under this specific low-speed scenario, capturing trajectories
rich in hysteresis, and (ii) Admittance control in the workspace typically
results in coupled joint motion, thereby coupled hysteresis. An NN model can
inherently discern these coupling patterns, which is often overlooked in
conventional model-based methods that assess joint frictions separately.

\subsection{Sensorless Hand-guiding}

\begin{figure}[t]
    \vspace{2pt}
    \centering
    \includegraphics[width=0.48\textwidth]{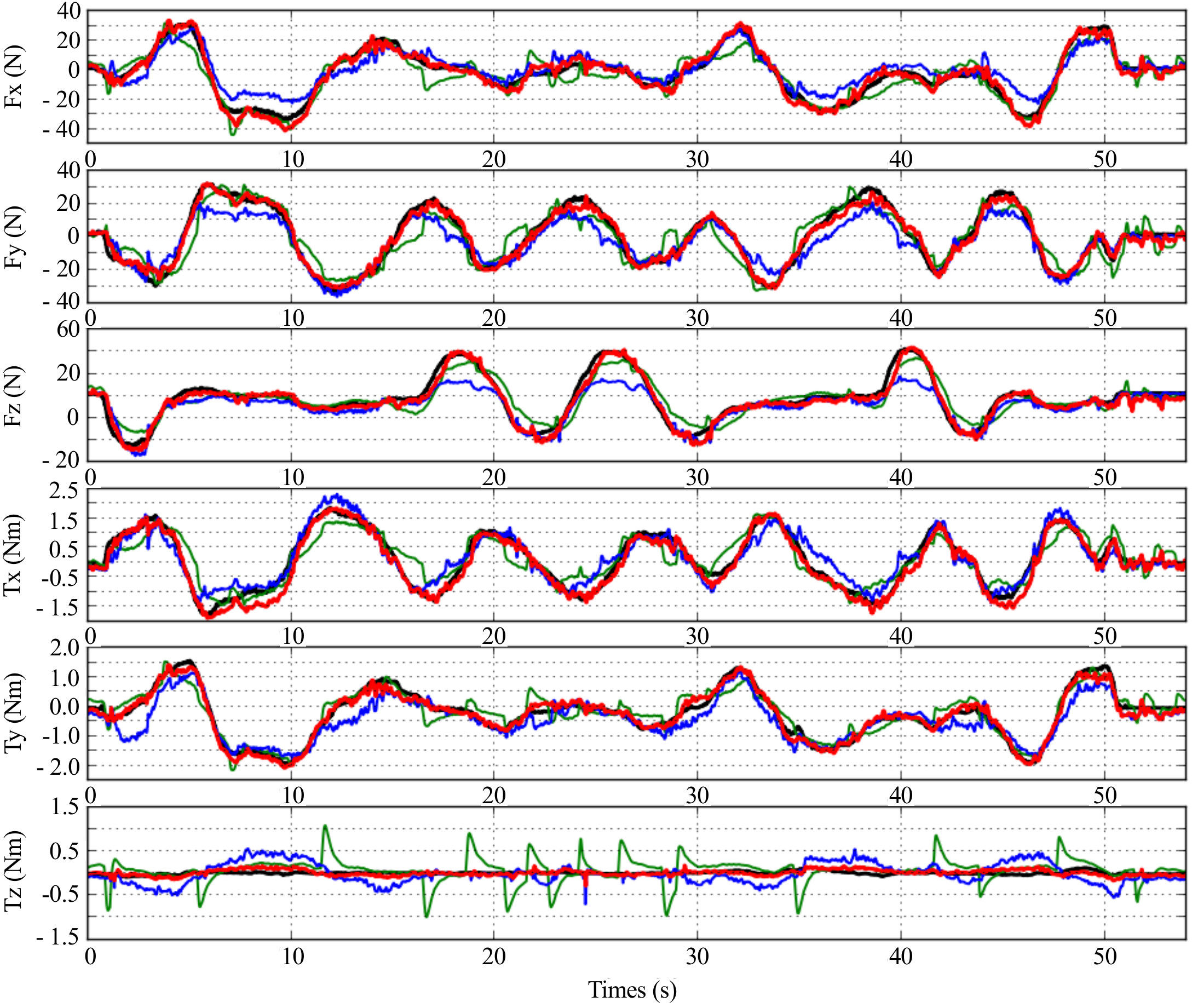}
    \vspace*{-5pt}
    \caption{Comparison between the fine-tuned model (red), base model (blue),
    and GMO (green) estimations with the measured wrench (black) in
    hand-guiding.}
    \label{fig:HG}
    \vspace*{-3pt}
\end{figure}

We employ the HGDS in this study to fine-tune the base model for enhanced
hand-guiding performance. Again, the HGDS is notably biased, as joint motions
consistently align with the end-effector forces due to the compliant nature of
the hand-guiding task. On the other hand, this straightforward motion pattern
elucidates why a smaller HGDS (comprising 321,000 data frames, in contrast to
the 991,500 frames in the FSDS) yields such a significant improvement in
accuracy upon fine-tuning the base model. This observation is evident in Fig.
\ref{fig:HG}, wherein both the base model and GMO occasionally miss peak values,
while the fine-tuned model consistently predicts values closely aligned with the
ground truth. A quantitative comparison is presented in Table
\ref{Table:HG_error}, suggesting that the estimation error is halved through the
fine-tuning process.

\subsection{Force/Torque Control for Tight Assembly}

In addition to the in-contact sliding task, the estimator's reliability was
further validated by successfully completing a widely-recognized industrial
task: tight assembly \cite{vuong2021learning, johannsmeier2019framework}.
Precise control over forces and torques in all directions is essential for pin
insertion. This task requires highly accurate force sensing with minimal time
delay.

In this experiment, a pair of aluminum pin and hole with diameters of 20mm and
0.1mm clearance were used. The hole was placed within the workspace for IK Class
1, and the pin was attached to the F/T sensor using a coupling. The NN-estimated
wrench was used for control instead the measurements. The insertion process
began 3mm above the hole, with a 2mm center and 5{\textdegree} orientation
misalignment. 

\begin{table}[t]
    \centering
    \setlength{\tabcolsep}{5.5pt}
    \begin{tabular}{c | c c c c c c}
    \hline
    \multicolumn{7}{c}{Fig. \ref{fig:HG} RMSE (N for F; Nm for T)} \\ \hline
    Model  & Fx & Fy & Fz & Tx & Ty & Tz \\ \hline
    CSDS Fine-tuned      & 3.29 & 3.07 & 4.25 & 0.22 & 0.15 & 0.07 \\
    FSDS Trained (base)  & 6.85 & 6.87 & 9.88 & 0.32 & 0.36 & 0.23 \\
    GMO                  & 6.11 & 7.28 & 8.63 & 0.42 & 0.29 & 0.28 \\ \hline
    \end{tabular}
    \caption{RMSE for the hand-guiding experiment.}
    \label{Table:HG_error}
    \vspace*{-5pt}
\end{table}

\begin{figure}[t]
    \vspace*{2pt}
    \centering
        \minipage{0.16\textwidth}
            \includegraphics[width=\linewidth]{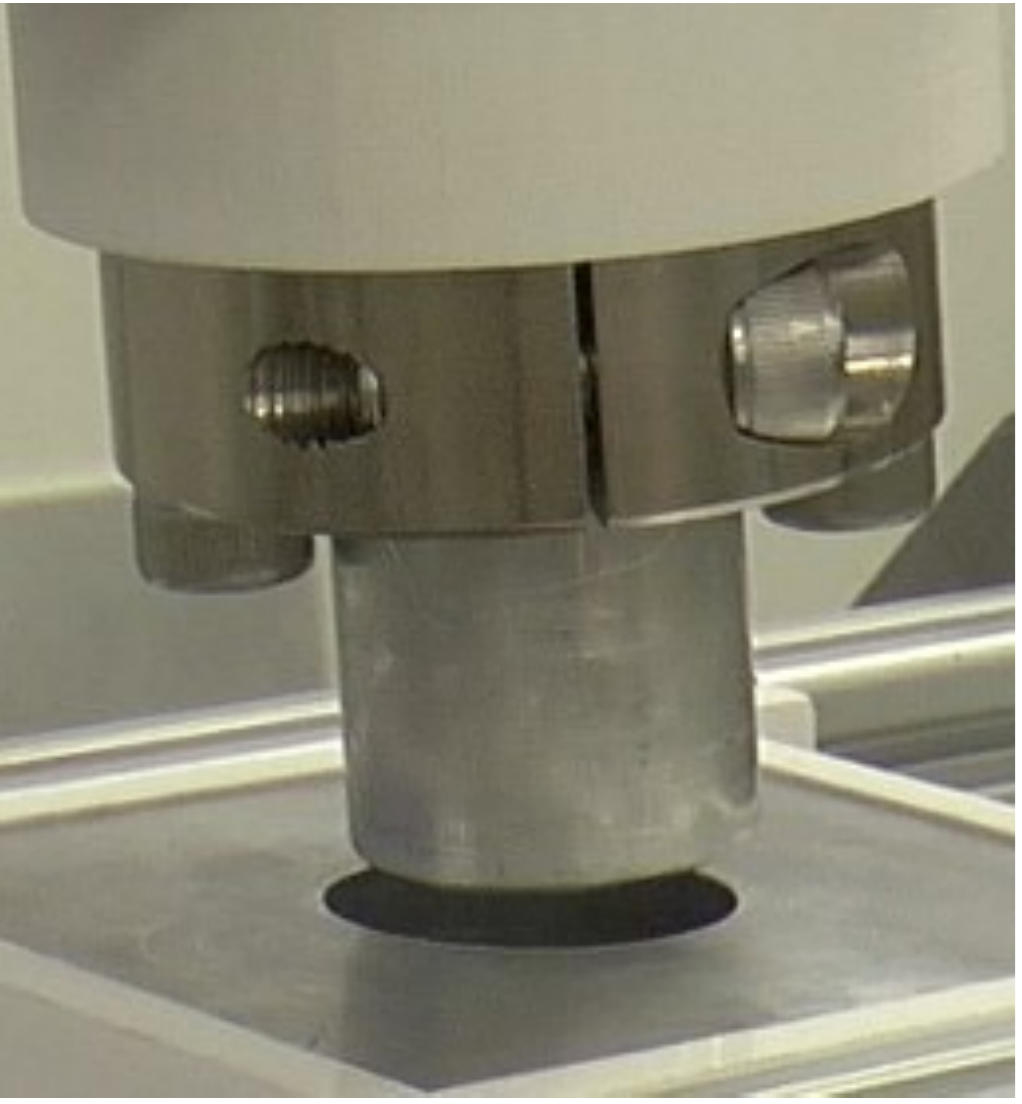}
        \endminipage\hfill
        \minipage{0.16\textwidth}
            \centering
            \includegraphics[width=\linewidth]{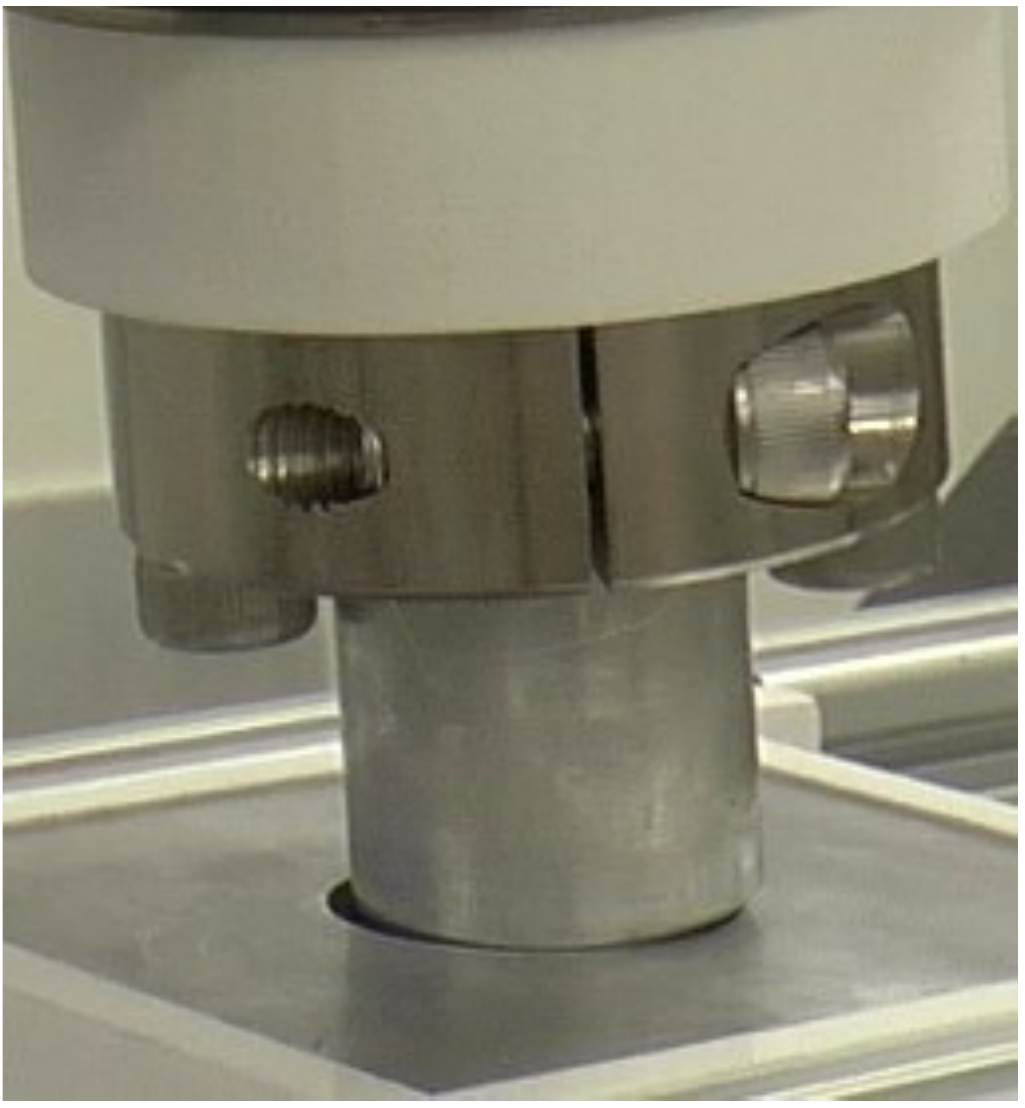}
        \endminipage\hfill
        \minipage{0.16\textwidth}
            \centering
            \includegraphics[width=\linewidth]{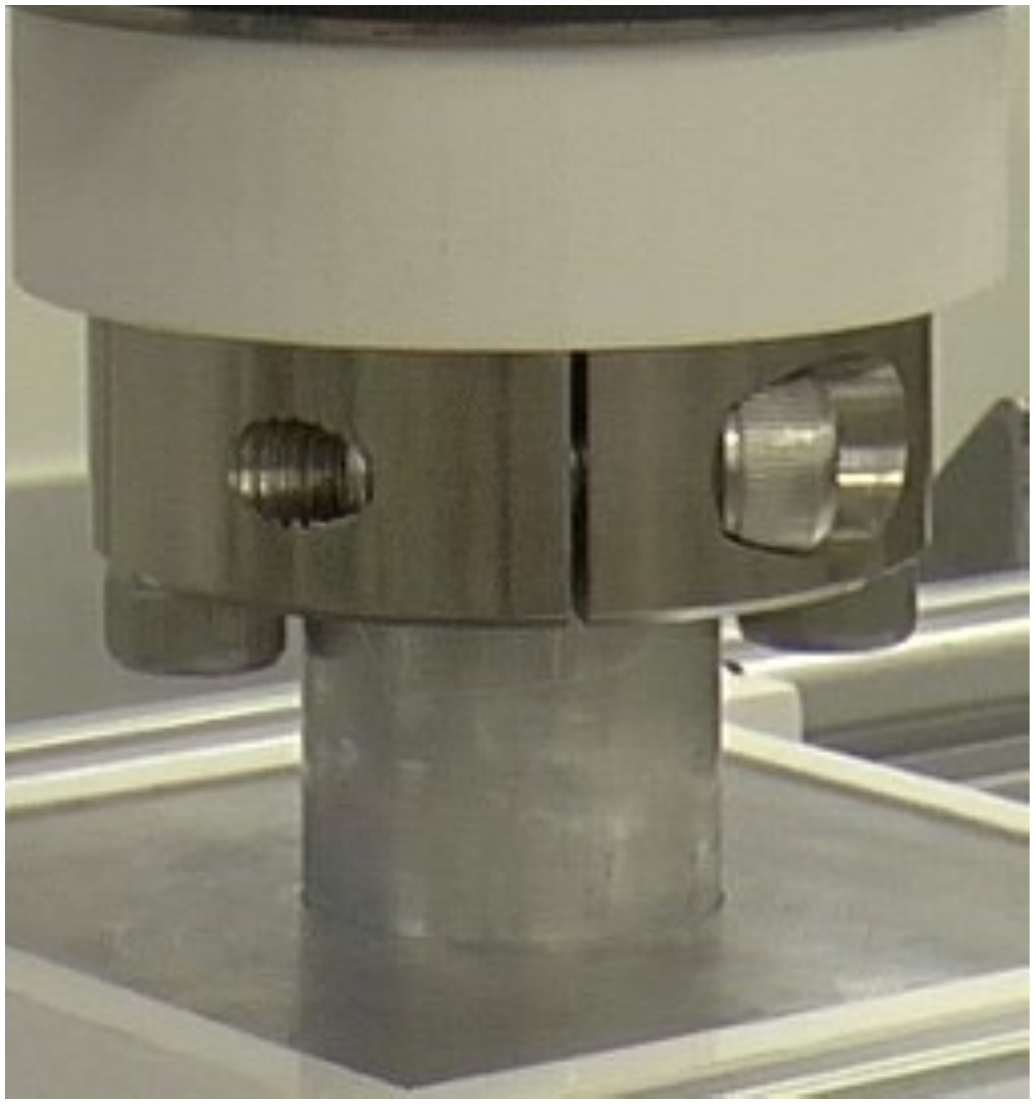}
        \endminipage\hfill \caption{Three-phase insertion procedure involving
        descending, aligning, and inserting.}
        \vspace*{-5pt}
        \label{Fig:PIH_stages}
\end{figure}

By setting the reference force in the pin-tip frame to [0N, 0N, 5N, 0Nm, 0Nm]
and leaving the end-effector twist uncontrolled for the cylinder pin, a
three-phase insertion process can be anticipated, depicted in Fig.
\ref{Fig:PIH_stages}. Initially, the pin moved downward until making contact.
Following this, due to the zero reference force and torque in the XY directions,
the pin automatically aligned both its center and orientation with the hole. In
the final phase, the pin continued its insertion, making fine adjustments to its
orientation. For this phase, another controller set with smaller gains was
employed, taking into account the tight clearance and rigid environment that
could result in high-magnitude force oscillations.

During the insertion phase, the pin made multi-point contact with the hole. This
presents a scenario where identical end-effector torques can result in different
torques at the joints, based on the number of contact points. For example, in a
single-point contact situation, the torques perceived at the end-effector get
amplified by the robot link when transmitted to the joints, leading to
significant current variations. Conversely, when there are two-point contacts in
opposite directions, no amplification occurs, resulting in only a subtle change
in the joint current measurement. Given the presence of noise, the NN model
often struggles to capture these slight variations, thus compromising accuracy
during the insertion phase. Such an issue is particularly challenging to
identify near certain robot configurations, such as singularities.

\begin{figure}[t]
    \centering
    \includegraphics[width=0.48\textwidth]{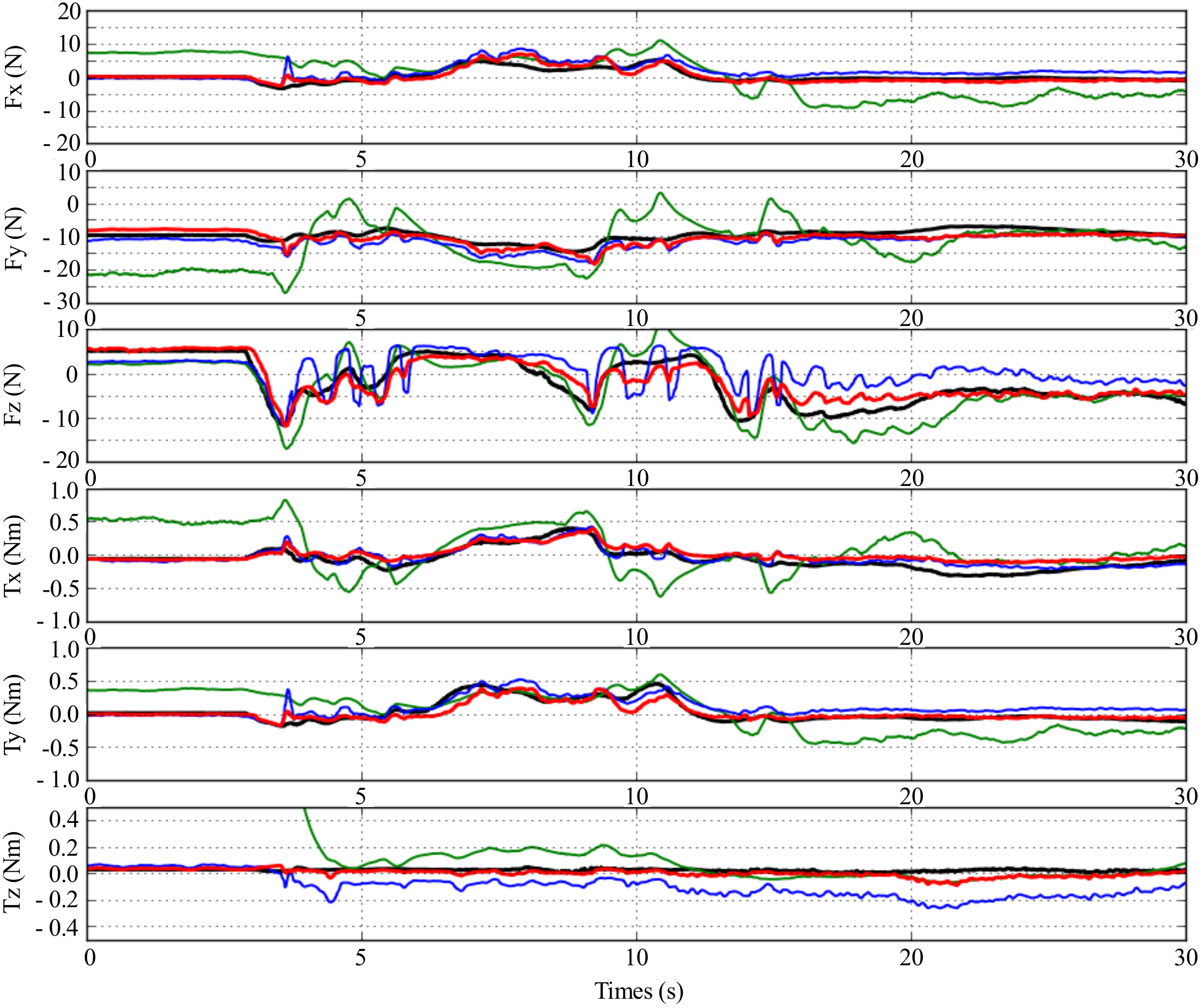}
    \vspace*{-5pt}
    \caption{Comparison between the fine-tuned model (red), base model (blue),
    and GMO (green) estimations with the measured wrench (black) in the pin
    insertion task.}
    \label{fig:PIH_wrench}
    \vspace*{-10pt}
\end{figure}

To tackle this problem, we collected additional training data by having the
robot execute the pin insertion task at various locations within the workspace.
This Pin Insertion DataSet (PIDS) contains 95,500 data frames. We haven't
addressed this dataset in Section IV since it was specifically designed for IK
Class 1 and confined to limited areas. Following this, we fine-tuned the model
using both CSDS and PIDS, which subsequently achieved an insertion depth of
18mm, as shown in Fig. \ref{fig:PIH_wrench}. Using only the CSDS to fine-tune
the base model would cause the pin to get stuck at 12mm. The CSDS+PIDS
fine-tuned model demonstrated reasonable accuracy during the initial 7 seconds
when the pin had single-point contact with the hole. However, it still presented
a relatively large error during the multi-contact phase due to the
aforementioned issue.

\begin{table}[t]
    \vspace{8pt}
    \centering
    \setlength{\tabcolsep}{5.5pt}
    \begin{tabular}{c | c c c c c c}
    \hline
    \multicolumn{7}{c}{Fig. \ref{fig:PIH_wrench} RMSE (N for F; Nm for T)} \\
    \hline
    Model  & Fx & Fy & Fz & Tx & Ty & Tz \\ \hline
    CSDS+PIDS Fine-tuned & 1.07 & 1.67 & 2.03 & 0.10 & 0.08 & 0.04 \\
    FSDS Trained (base)  & 1.94 & 2.15 & 4.56 & 0.08 & 0.11 & 0.15 \\
    GMO                  & 5.53 & 6.75 & 3.19 & 0.34 & 0.25 & 0.39 \\ \hline
    \end{tabular}
    \caption{RMSE for the pin insertion experiment.}
    \label{Table:PIH_error}
    \vspace*{-8pt}
\end{table}

\section{Conclusion}

In this paper, we propose an approach to estimating the end-effector wrench with
Neural Networks. The model takes the joint currents and states as input and
estimates the external wrench in real-time. It avoids using embedded joint
torque sensors and can replace 6-axis end-effector F/T sensors in industrial
applications. The implemented model has demonstrated high estimation accuracy
and stability across different industrial tasks, being trained with the
well-categorized datasets and enhanced with the fine-tuning strategy. 

All training datasets were collected under low-speed conditions, consistent with
the contact-rich tasks demonstrated in this paper. Future research might explore
potential challenges, including safety concerns, associated with data collection
during high-speed operations. Moreover, devising a calibration scheme could
enhance the method's adaptability when changing the end-effector or attaching
supplementary equipment to the robot. Another promising avenue for future
research is to combine the sensorless F/T estimation here with recent robust
force control methods \cite{pham2020convex}, which can further alleviate
possible instabilities caused by estimation errors or delays. Such a combination
could enable even more sensitive sensorless manipulation.
\vspace*{-5pt}

\section{Acknowledgement}

This research was supported by the National Research Foundation, Prime
Minister's Office, Singapore under its Medium Sized Centre funding scheme,
Singapore Centre for 3D Printing, CES\_SDC Pte Ltd, and Chip Eng Seng
Corporation Ltd.

\bibliographystyle{ieeetr}
\bibliography{reference} 

\begin{thebibliography}{10}

\bibitem{siciliano2008springer}
B.~Siciliano, O.~Khatib, and T.~Kr{\"o}ger, {\em Springer handbook of
  robotics}, vol.~200.
\newblock Springer, 2008.

\bibitem{suarez2018can}
F.~Su{\'a}rez-Ruiz, X.~Zhou, and Q.-C. Pham, ``Can robots assemble an ikea
  chair?,'' {\em Science Robotics}, vol.~3, no.~17, p.~eaat6385, 2018.

\bibitem{pham2020convex}
H.~Pham and Q.-C. Pham, ``Convex controller synthesis for robot contact,'' {\em
  IEEE Robotics and Automation Letters}, vol.~5, no.~2, pp.~3330--3337, 2020.

\bibitem{de2005sensorless}
A.~De~Luca and R.~Mattone, ``Sensorless robot collision detection and hybrid
  force/motion control,'' in {\em Proceedings of the 2005 IEEE international
  conference on robotics and automation}, pp.~999--1004, IEEE, 2005.

\bibitem{van2011estimating}
M.~Van~Damme, P.~Beyl, B.~Vanderborght, V.~Grosu, R.~Van~Ham, I.~Vanderniepen,
  A.~Matthys, and D.~Lefeber, ``Estimating robot end-effector force from noisy
  actuator torque measurements,'' in {\em 2011 IEEE International Conference on
  Robotics and Automation}, pp.~1108--1113, IEEE, 2011.

\bibitem{wahrburg2017motor}
A.~Wahrburg, J.~B{\"o}s, K.~D. Listmann, F.~Dai, B.~Matthias, and H.~Ding,
  ``Motor-current-based estimation of cartesian contact forces and torques for
  robotic manipulators and its application to force control,'' {\em IEEE
  Transactions on Automation Science and Engineering}, vol.~15, no.~2,
  pp.~879--886, 2017.

\bibitem{9382115}
S.~K. Kommuri, S.~Han, and S.~Lee, ``External torque estimation using higher
  order sliding-mode observer for robot manipulators,'' {\em IEEE/ASME
  Transactions on Mechatronics}, vol.~27, no.~1, pp.~513--523, 2022.

\bibitem{nguyen2009model}
D.~Nguyen-Tuong, M.~Seeger, and J.~Peters, ``Model learning with local gaussian
  process regression,'' {\em Advanced Robotics}, vol.~23, no.~15,
  pp.~2015--2034, 2009.

\bibitem{liu2018end}
X.~Liu, F.~Zhao, S.~S. Ge, Y.~Wu, and X.~Mei, ``End-effector force estimation
  for flexible-joint robots with global friction approximation using neural
  networks,'' {\em IEEE Transactions on Industrial Informatics}, vol.~15,
  no.~3, pp.~1730--1741, 2018.

\bibitem{sharkawy2020human}
A.-N. Sharkawy, P.~N. Koustoumpardis, and N.~Aspragathos, ``Human--robot
  collisions detection for safe human--robot interaction using one
  multi-input--output neural network,'' {\em Soft Computing}, vol.~24, no.~9,
  pp.~6687--6719, 2020.

\bibitem{kim2021transferable}
D.~Kim, D.~Lim, and J.~Park, ``Transferable collision detection learning for
  collaborative manipulator using versatile modularized neural network,'' {\em
  IEEE Transactions on Robotics}, 2021.

\bibitem{khalil2002modeling}
W.~Khalil and E.~Dombre, {\em Modeling identification and control of robots}.
\newblock CRC Press, 2002.

\bibitem{de2003actuator}
A.~De~Luca and R.~Mattone, ``Actuator failure detection and isolation using
  generalized momenta,'' in {\em 2003 IEEE international conference on robotics
  and automation (cat. No. 03CH37422)}, vol.~1, pp.~634--639, IEEE, 2003.

\bibitem{liu2015experimental}
Y.~Liu, J.~Li, Z.~Zhang, X.~Hu, and W.~Zhang, ``Experimental comparison of five
  friction models on the same test-bed of the micro stick-slip motion system,''
  {\em Mechanical Sciences}, vol.~6, no.~1, pp.~15--28, 2015.

\bibitem{liu2021sensorless}
S.~Liu, L.~Wang, and X.~V. Wang, ``Sensorless haptic control for human-robot
  collaborative assembly,'' {\em CIRP Journal of Manufacturing Science and
  Technology}, vol.~32, pp.~132--144, 2021.

\bibitem{gaz2018model}
C.~Gaz, E.~Magrini, and A.~De~Luca, ``A model-based residual approach for
  human-robot collaboration during manual polishing operations,'' {\em
  Mechatronics}, vol.~55, pp.~234--247, 2018.

\bibitem{gaz2019dynamic}
C.~Gaz, M.~Cognetti, A.~Oliva, P.~R. Giordano, and A.~De~Luca, ``Dynamic
  identification of the franka emika panda robot with retrieval of feasible
  parameters using penalty-based optimization,'' {\em IEEE Robotics and
  Automation Letters}, vol.~4, no.~4, pp.~4147--4154, 2019.

\bibitem{hu2017contact}
J.~Hu and R.~Xiong, ``Contact force estimation for robot manipulator using
  semiparametric model and disturbance kalman filter,'' {\em IEEE Transactions
  on Industrial Electronics}, vol.~65, no.~4, pp.~3365--3375, 2017.

\bibitem{schaal2002scalable}
S.~Schaal, C.~G. Atkeson, and S.~Vijayakumar, ``Scalable techniques from
  nonparametric statistics for real time robot learning,'' {\em Applied
  Intelligence}, vol.~17, no.~1, pp.~49--60, 2002.

\bibitem{vijayakumar1997local}
S.~Vijayakumar and S.~Schaal, ``Local dimensionality reduction for locally
  weighted learning,'' in {\em Proceedings 1997 IEEE International Symposium on
  Computational Intelligence in Robotics and Automation CIRA'97.'Towards New
  Computational Principles for Robotics and Automation'}, pp.~220--225, IEEE,
  1997.

\bibitem{nguyen2008local}
D.~Nguyen-Tuong and J.~Peters, ``Local gaussian process regression for
  real-time model-based robot control,'' in {\em 2008 IEEE/RSJ International
  Conference on Intelligent Robots and Systems}, pp.~380--385, IEEE, 2008.

\bibitem{gijsberts2013real}
A.~Gijsberts and G.~Metta, ``Real-time model learning using incremental sparse
  spectrum gaussian process regression,'' {\em Neural networks}, vol.~41,
  pp.~59--69, 2013.

\bibitem{yilmaz2020neural}
N.~Yilmaz, J.~Y. Wu, P.~Kazanzides, and U.~Tumerdem, ``Neural network based
  inverse dynamics identification and external force estimation on the da vinci
  research kit,'' in {\em 2020 IEEE International Conference on Robotics and
  Automation (ICRA)}, pp.~1387--1393, IEEE, 2020.

\bibitem{lee2018interaction}
D.-H. Lee, W.~Hwang, and S.-C. Lim, ``Interaction force estimation using camera
  and electrical current without force/torque sensor,'' {\em IEEE Sensors
  Journal}, vol.~18, no.~21, pp.~8863--8872, 2018.

\bibitem{xia2021sensorless}
J.~Xia and K.~Kiguchi, ``Sensorless real-time force estimation in microsurgery
  robots using a time series convolutional neural network,'' {\em IEEE Access},
  vol.~9, pp.~149447--149455, 2021.

\bibitem{farazi2017online}
H.~Farazi and S.~Behnke, ``Online visual robot tracking and identification
  using deep lstm networks,'' in {\em 2017 IEEE/RSJ International Conference on
  Intelligent Robots and Systems (IROS)}, pp.~6118--6125, IEEE, 2017.

\bibitem{wu2021hysteresis}
D.~Wu, Y.~Zhang, M.~Ourak, K.~Niu, J.~Dankelman, and E.~Vander~Poorten,
  ``Hysteresis modeling of robotic catheters based on long short-term memory
  network for improved environment reconstruction,'' {\em IEEE Robotics and
  Automation Letters}, vol.~6, no.~2, pp.~2106--2113, 2021.

\bibitem{goodfellow2016deep}
I.~Goodfellow, Y.~Bengio, and A.~Courville, {\em Deep learning}.
\newblock MIT press, 2016.

\bibitem{hochreiter1997long}
S.~Hochreiter and J.~Schmidhuber, ``Long short-term memory,'' {\em Neural
  computation}, vol.~9, no.~8, pp.~1735--1780, 1997.

\bibitem{zhang2022robot}
T.~Zhang, X.~Liang, and Y.~Zou, ``Robot peg-in-hole assembly based on contact
  force estimation compensated by convolutional neural network,'' {\em Control
  Engineering Practice}, vol.~120, p.~105012, 2022.

\bibitem{lecun1995convolutional}
Y.~LeCun and Y.~Bengio, ``Convolutional networks for images, speech, and time
  series,'' {\em The handbook of brain theory and neural networks}, vol.~3361,
  no.~10, p.~1995, 1995.

\bibitem{vuong2021learning}
N.~Vuong, H.~Pham, and Q.-C. Pham, ``Learning sequences of manipulation
  primitives for robotic assembly,'' in {\em 2021 IEEE International Conference
  on Robotics and Automation (ICRA)}, pp.~4086--4092, IEEE, 2021.

\bibitem{johannsmeier2019framework}
L.~Johannsmeier, M.~Gerchow, and S.~Haddadin, ``A framework for robot
  manipulation: Skill formalism, meta learning and adaptive control,'' in {\em
  2019 International Conference on Robotics and Automation (ICRA)},
  pp.~5844--5850, IEEE, 2019.

\end{thebibliography}

\end{document}